\documentclass[a4paper,fleqn]{cas-dc}

\usepackage[authoryear,longnamesfirst]{natbib}
\usepackage{amsmath}
\usepackage{amssymb}
\usepackage{mathtools}
\usepackage{amsthm}
\usepackage{pifont}
\newcommand{\cmark}{\ding{51}}%
\newcommand{\xmark}{\ding{55}}%
\usepackage{bm}
\usepackage{placeins}
\usepackage{afterpage}
\usepackage{paralist}
\usepackage[inline]{enumitem}
\usepackage{graphicx}
\usepackage{subcaption}
\usepackage{microtype}

\def\tsc#1{\csdef{#1}{\textsc{\lowercase{#1}}\xspace}}
\tsc{WGM}
\tsc{QE}

\begin{document}
\let\WriteBookmarks\relax
\def\floatpagepagefraction{1}
\def\textpagefraction{.001}

\shorttitle{}    

\shortauthors{}  

\title [mode = title]{ParamBoost: Gradient Boosted Piecewise Cubic Polynomials}  



%

\author[1]{Nicolas Salvad\'e}[orcid=0009-0000-0961-9844]



\ead{nicolas.salvade.22@ucl.ac.uk}


\credit{Writing – original draft, Visualization, Software, Methodology, Conceptualization, Formal analysis, Investigation}

\affiliation[1]{organization={Department of Civil, Environmental and Geomatic Engineering, University College London},
            addressline={Gower Street}, 
            city={London},
            postcode={WC1E 6BT}, 
            state={},
            country={United Kingdom}}

\author[1]{Tim Hillel}[orcid=0000-0001-6872-2235]


\cormark[2]

\ead{tim.hillel@ucl.ac.uk}


\credit{Writing – review \& editing, Supervision, Methodology, Conceptualization, Project administration}


\cortext[2]{Corresponding author}



\begin{abstract}
Generalized Additive Models (GAMs) can be used to create non-linear \emph{glass-box} (i.e. explicitly interpretable) models, where the predictive function is fully observable over the complete input space. However, glass-box interpretability itself does not allow for the incorporation of expert knowledge from the modeller. In this paper, we present \textbf{ParamBoost}, a novel GAM whose shape functions (i.e. mappings from individual input features to the output) are learnt using a Gradient Boosting algorithm that fits cubic polynomial functions at leaf nodes. ParamBoost incorporates several constraints commonly used in parametric analysis to ensure well-refined shape functions. These constraints include:
\begin{enumerate*}[label=(\roman*)]
    \item continuity of the shape functions and their derivatives (up to $C^2$);
    \item monotonicity;
    \item convexity;
    \item feature interaction constraints; and 
    \item model specification constraints. 
\end{enumerate*}
Empirical results show that the unconstrained ParamBoost model consistently outperforms state-of-the-art GAMs across several real-world datasets. We further demonstrate that modellers can selectively impose required constraints at a modest trade-off in predictive performance, allowing the model to be fully tailored to application-specific interpretability and parametric-analysis requirements.
\end{abstract}


\begin{highlights}
\item ParamBoost is a new GAM that uses Gradient Boosting with cubic polynomials to learn shape functions.
\item Existing glass-box models lack support for incorporating expert knowledge.
\item It supports constraints like monotonicity, convexity, and smooth continuity.
\item The unconstrained model outperforms state-of-the-art GAMs on real-world datasets.
\item Constraints can be selectively applied with only a modest trade-off in accuracy.
\end{highlights}


\begin{keywords}
Machine Learning \sep Gradient Boosting \sep Generalized Additive Models \sep Interpretability \sep Shape Function Constraints
\end{keywords}

\maketitle

\section{Introduction}

Machine Learning (ML) methods (and Deep Neural Networks (DNNs) in particular) offer improved predictive capabilities over traditional parametric models. However, many applications still require parameter-based interpretations, where practitioners rely on gradients (i.e., \emph{marginal effects} or \emph{elasticities}) to quantify how a feature (or \emph{exogenous variable}) influences the output. Such analysis is crucial in domains including:
\begin{enumerate*}[label=(\roman*)]
    \item transportation modelling \citep{wardman2022meta, salvade2025rumboost};
    \item econometrics \citep{zhang2022measurement};
    \item health \citep{zazdravnykh2024total}; and
    \item structural equation modelling and energy demand \citep{gao2021income}.
\end{enumerate*}

Recent years have seen a growing focus on \textit{interpretable} ML, leading to the establishment of the field of eXplainable Artificial Intelligence (XAI). XAI techniques largely fall into two different approaches:
\begin{enumerate*}[label=(\roman*)]
    \item \emph{post-hoc} techniques that approximate a black-box model; and
    \item \emph{intrinsically interpretable} glass- (or white-) box models. 
\end{enumerate*}
Among post-hoc methods, permutation-based techniques such as
LIME \citep{ribeiro2016should} and SHAP \citep{lundberg2017consistent}, have been most widely used, with applications in healthcare \citep{gabbay2021lime, zhang2022machine, magesh2020explainable}, finance \citep{gramegna2021shap, lin2022model, aljadani2023mathematical}, road safety \citep{yang2021application, parsa2020toward} and travel behaviour \citep{TAMIMKASHIFI2022279, dahmen2023interpretable, Martin-Baos2023-xb, ren2023exploring, HAGENAUER2017273, SALAS2022116253}. 
However, as these methods rely on local approximations of the model \textit{predictive} functions, they only provide valid explanations at observed data points and offer no guarantee of validity over the full input space. 

On the other hand, recent work in intrinsically interpretable techniques has focused on Generalised Additive Models (GAMs)\citep{hastie1990generalized}, where each feature is associated with a non-linear shape function learnt from the data; commonly using splines or Gaussian processes (GPs) \citep{wood2017generalized}, and more recently incorporating structural constraints \citep{ibrahim2025predicting}). Other ML algorithms have also been adapted to GAMs, including scalable tree-based methods \citep{caruana2015intelligible, konstantinov2021interpretable, chang2021node, ibrahim2023grand, gabidollageneralized, salvade2025rumboost}, NNs \citep{agarwal2021neural, radenovic2022neural, yang2021gami}, piecewise-linear models \citep{von2024automatic} and polynomial-based models \citep{dubey2022scalable, duong2024cat}, often achieving performance close to black-box models. 

Despite these advances, glass-box interpretability itself does not enable the incorporation of domain knowledge. We identify four essential constraints that, when applied to a GAM, allow for the integration of expert knowledge: 
\begin{enumerate*}[label=(\roman*)]
    \item \emph{continuous differentiability} of shape functions over the full input space, enabling computation of marginal effects and related indicators;
    \item \emph{restrictable feature interactions}, enabling each feature's effect on the predictive function to be represented and visualised with a \textit{single} exact shape function;
    \item \emph{monotonicity or curvature (i.e. convex\slash concave) control} of shape functions; and
    \item \emph{output-specific feature assignment} in multi-output problems (e.g. multi-class and multi-label classification, multi-target regression, and other multitask problems), such that only relevant features influence each predictive function. For example, for disease prediction in blood tests, indicators related to a specific disease should only appear in the corresponding function (e.g. a white blood cell count should influence the blood cancer class only).
\end{enumerate*}

While some existing methods satisfy some individual constraints above, none provide all four at the same time, limiting their usage in domains such as health, transport, and policy evaluation. Table \ref{model comparison} summarises the interpretability capabilities of state-of-the-art GAMs.

\begin{table*}[ht]

\tiny
\centering
\caption{Comparison of shape function constraint capabilities of state-of-the-art GAMs. \textbf{Diff.:} continuous differentiability of shape functions; \textbf{FI:} restrictable feature interactions; \textbf{Mono.\slash curv.} monotonicity and\slash or curvature control of shape functions; \textbf{OSF:} output-specific feature assignment in multi-output problems}
\begin{tabular}{lllcccc}
\toprule
\textbf{Model} & \textbf{Modelling tasks} & \textbf{Typical base learner} & \textbf{Diff.} & \textbf{FI} & \textbf{Mono.\slash curv.} & \textbf{OSF}\\
\midrule
EBM \citep{caruana2015intelligible, chang2021node} & Regr. and class. & Trees & \xmark & \cmark & $\sim$* & \xmark\\
EGBM \citep{konstantinov2021interpretable} & Regr. & Randomised trees  & \xmark & \xmark & \xmark & \xmark\\
Stump Forests \citep{gabidollageneralized} & Regr. and class. & Trees (stumps) & \xmark & \cmark & \xmark & \xmark\\
GRAND-SLAMIN \citep{ibrahim2023grand} & Regr. and class. & Soft trees & $C^{0}$ & \cmark & \xmark & \xmark\\
APLR \citep{von2024automatic} & Regr. and class. & Linear trees & $C^{0}$ & \cmark & $\sim$* & \xmark \\
NAM \citep{agarwal2021neural} & Regr. and class. & Neural networks & \xmark & \cmark & \xmark & \xmark \\
GAMI-Net \citep{yang2021gami} & Regr. and class.** & Neural networks & \xmark & \cmark & \xmark*** & \xmark\\
NBM \citep{radenovic2022neural} & Regr. and class. & Neural basis & \xmark & \cmark & \xmark & \xmark \\
SPAM \citep{dubey2022scalable} & Regr. and class. & Polynomials & \xmark & \xmark**** & \xmark & \xmark \\
CAT \citep{duong2024cat} & Regr. and class. & Polynomials & \xmark & \xmark & \xmark & \xmark \\
MGCV \citep{wood2017generalized} & Regr. and class. & Splines & $C^{2}$***** & \cmark & \xmark & \cmark\\
ELAAN \citep{ibrahim2025predicting} & Regr. & Splines & $C^{2}$***** & \cmark & \xmark & \xmark \\
\midrule
ParamBoost (our approach) & Regr. and class. & Cubic trees & $C^{2}$ & \cmark & \cmark & \cmark \\
\bottomrule
\multicolumn{7}{r}{*only monotonicity constraints}\\
\multicolumn{7}{r}{**only binary classification}\\
\multicolumn{7}{r}{***based on the paper}\\
\multicolumn{7}{r}{****if interactions are turned off, the model falls back to linear\slash logistic regression}\\
\multicolumn{7}{r}{*****with cubic splines}\\
\label{model comparison}
\end{tabular}

\label{tab:model_comparison}
\end{table*}

To address these limitations, we propose \textbf{ParamBoost}, a novel parametric gradient boosting GAM that fits cubic polynomial functions at leaf nodes. This allows each shape function to be expressed as a polynomial up to cubic degree, enabling flexible non-linear effects while maintaining continuous gradients for computing marginal effects and confidence intervals. ParamBoost introduces four tunable model components:
\begin{enumerate}
    \item \textbf{Degree $D$ of the polynomial trees $O^D$:} up to the cubic degree ($O^3$), where $D=0$ reduces to classical Gradient Boosting Decision Trees (GBDT);
    \item \textbf{Smoothness $S$ of shape functions $C^S$:}, from discontinuous ($S=-1$) to $C^2$ (i.e. continuous second derivative), subject to $S\leq D-1$;
    \item \textbf{Monotonicity and curvature constraints:} restricting each feature's shape function to incorporate domain knowledge; and
    \item \textbf{Output-specific feature assignment:} restricting which features impact which predictive functions, again incorporating domain knowledge and supporting meaningful shape constraints\footnote{as noted by \citet{zhang2019axiomatic}, if a feature is present in all class-specific predictive functions, monotonicity in the shape function does not guarantee monotonicity of the probability outputs due to the softmax. Restricting a feature to a single predictive function resolves this, with the same reasoning applied to curvature constraints.}.
\end{enumerate}

ParamBoost incorporates these constraints during training, which has several advantages:
\begin{enumerate*}[label=(\roman*)]
    \item the shape functions are guaranteed to behave in a way that respects domain knowledge;
    \item the model can find the best solution within the constrained space, and there are no loss of predictive power from a post-hoc transformation, that would happen, for example, when there is a major violation of the constraint; and
    \item constraints can act as a form of regularisation.
\end{enumerate*}

We evaluate ParamBoost on 11 real-world datasets across regression, binary classification, and multi-class classification tasks. We then present a detailed case study to demonstrate how the proposed constraints improve interpretability and generalisation. The code used to run experiments is available on Github\footnote{\url{https://github.com/big-ucl/paramboost}}.

\section{Methodology} \label{methodology}

\subsection{GAMs} \label{gams}

GAMs are additive models where the predictive function $F_i$ for a class or task $i$ is the sum of specific shape functions $f_{i,k}$:
\begin{equation}
    F_i(\bm{x}) = \beta_0 + \sum_{k=1}^{K} f_{i,k}(x_k),
\end{equation}
where $x_k$ is the data associated with feature $k$. These predictive functions are transformed into predictions $\hat{y}$, the inverse of the link function $g$. The type of link function used depends on the nature of the problem: (i)~for (multi-target) regression, the identity function:
\begin{equation}
    \hat{y}_i = F_i(\bm{x}) \quad \forall i= 1,..., J,
\end{equation}
where $J=1$ for a simple regression task;
(ii)~for multi-label or binary classification, the logistic function:
\begin{equation}
    \hat{y_i} = \frac{1}{1+e^{F_i(\bm{x})}} \quad \forall i = 1, ..., J,
\end{equation}
where $J=1$ for binary classification; and
(iii)~for multi-class classification, the softmax function:
\begin{equation}
    \hat{y}_i = \frac{e^{F_i(\bm{x})}}{\sum_{j=1}^{J}e^{F_j(\bm{x})}},
\end{equation}
where $J$ is the number of classes.

Each shape function is learnt by minimising a loss function $\mathcal{L}$ over the dataset of size $N$:
\begin{equation} \label{loss}
    \mathcal{L} = \frac{1}{N}\sum_n\sum_i\ell(y_{in},\hat{y}_{in}),
\end{equation}
where $\ell$ is the element-wise loss, 
typically:
\begin{enumerate*}[label=(\roman*)]
    \item the mean squared error $(y_{in} - \hat{y}_{in})^2$ for regression and
    \item the cross-entropy loss $\sum_i y_{in}\ln(\hat{y}_{in})$ for classification.
\end{enumerate*}
Note that $y_{in}$ is the observed target for class $i$ and observation $n$ (continuous for regression; equals to one if $i$ is the outcome, zero otherwise for classification).

\subsection{Gradient Boosting Decision Trees}

Gradient Boosting Decision Trees (GBDT) constructs ensembles of regression trees, where each tree partitions the input space and assigns a constant leaf-value to each partition. At iteration $m$, a new tree is added to the ensemble of class\slash task $i$ to minimise the loss function directly:

\begin{equation}
    \mathcal{L} = \sum_{n=1}^{N}\sum_{i=1}^{J}\ell\left(y_{in}, \hat{y}_{in, m-1} + T_{i,m}(x_{n})\right),
\end{equation}
where:
\begin{enumerate*}[label=(\roman*)]
    \item $\ell$ is the loss function for each individual and alternative as defined in Equation~\ref{loss};
    \item $\hat{y}_{in, m-1} = \sum_{p=1}^{m-1} T_{i,p}(x_{n})$ is the current prediction of the ensemble for observation $n$ and class\slash task $i$; and
    \item $T_{i,m}(x_{n})$ is the tree added at iteration $m$.
\end{enumerate*}

To determine the optimal leaf values of the new tree, GBDT uses a second-order Taylor expansion of the loss function around the current prediction:
\begin{equation}\label{newton_raphston}
\begin{aligned}
    \mathcal{L} \approx \sum_{n=1}^{N}\sum_{i=1}^{J} \big[ \ell(&y_{in},\hat{y}_{in, m-1}) \\&+g_{in} T_{i,m}({x}_{n}) + \frac{1}{2} h_{in} T_{i,m}({x}_{n})^2 \big],
\end{aligned}
\end{equation}
where $g_{in} = \partial\ell(y_{in}, \hat{y}_{in, m-1})/\partial\hat{y}_{in, m-
1}$ and $h_{in} = \partial^2\ell(y_{in},$ $ \hat{y}_{in, m-1})/
\partial^2\hat{y}_{in, m-1}$ are the first and second derivatives, respectively, of the loss with respect to the prediction.

In standard GBDT, each tree $T_{i,m}(x_{n})$ consists of $L$ non-overlapping terminal nodes (or \emph{leaves}) such that each data point will belong to a single leaf:
\begin{equation}
    T_{i,m}(x_{n}) = \sum_{l=1}^{L} \gamma_{l,i,m}  \mathbf{1}(x_{n}\in l),
\end{equation}
where $\mathbf{1}(x_{n}\in l)$ indicates that sample $x_n$ belongs to leaf $l$.

The optimal value $\gamma_{l,i,m}$ of a leaf~$l$ of class\slash task~$i$ at iteration~$m$ is obtained by setting the derivative of Equation \ref{newton_raphston} with respect to the leaf value to zero, giving:
\begin{equation}\label{leaf}
    \gamma_{l,i,m} = - \frac{(\sum_{n \in l}g_{in})}{(\sum_{n \in l}h_{in})}
\end{equation}

GBDT therefore effectively mimics a mini-batch \textit{quasi-Newton} method with a diagonal approximation of the Hessian applied to a batch defined by the tree partitions.

\subsection{Gradient Boosting with Polynomial Trees}

We modify the standard paradigm of GBDT by assuming that the terminal nodes output \emph{monomials} of degree $d \in \{0,..,D\}$, rather than constant values. Thus, the tree predictions $f_{i,m} (x_{in})$ correspond to elements of a polynomial function rather than shape function values. Gradient boosting with linear trees (monomial of degree 1) was introduced by \citet{shi_gradient_2019}, but their approach fits \emph{independent} linear models $\beta_{i0} + \boldsymbol{\beta}_{i}^{T} \mathbf{x_{n}} $ on the left and right sides of each split, providing no mechanism to enforce continuity. In contrast, we exploit the binning procedure of histogram-based GBDT, where each feature is discretised into $B$ bins. For each feature $k$ and bin $b \in \{1, ...,B\},$ the binned variable is defined as
\begin{equation}
    x^*_{kbn} =
    \begin{cases} 
    0 & \text{if } x_{kn} < u_{b-1}, \\
    x_{kn} &\text{if } x_{kn} < u_{b} \text{ and }  b=1\\
    x_{kn} - u_{b-1} & \text{if } u_{b-1} \leq x_{kn} < u_{b} \text{ and } b>1, \\ 
    u_b & \text{otherwise},
    \end{cases}
    \quad\quad
\end{equation}
where $u_b$ denotes the upper boundary of the bin (and $u_0=-\infty$ and $u_B=\infty$). Note that the bin edges are obtained with a quantile-based binning algorithm. Thus, each feature shape function becomes a piecewise polynomial of degree $D$:
\begin{equation}
    f_{ik}(x_{kn}) =  \sum_{d=0}^{D} \sum_{b=1}^{B}\beta_{ikdb} (x^*_{kbn})^{d}.
\end{equation}

As before, we apply the second-order approximation in \ref{newton_raphston} to find the optimal parameter values, however the gradients and Hessians must now account for the polynomial basis. Using the chain rule, the first derivative of the loss with respect to coefficient $\beta_{ikdb}$ is given by:
\begin{equation}
\begin{split}
    g^{\beta}_{ikbdn} & = \frac{\partial\ell(y_{in}, \hat{y}_{in, m-1})}{\partial\beta_{ikdb}}\\ 
    &= g_{in} \frac{\partial F_{in, m-1}}{\partial\beta_{ikdb}} = g_{in} (x^*_{kbn})^{d}.
\end{split}
\end{equation}

Similarly, the second derivative is given by:
\begin{equation}\label{hessian}
\begin{split}
    h^{\beta}_{ikbdn} & = \frac{\partial^2\ell(y_{in}, \hat{y}_{in, m-1})}{\partial^2\beta_{ikdb}} \\
    & = h_{in}\left(\frac{\partial F_{in, m-1}}{\partial\beta_{ikdb}}\right)^2 
     = h_{in} (x^*_{kbn})^{2d},
\end{split}
\end{equation}

The optimal leaf values in each terminal node~$l$ are then obtained as with in Equation \ref{leaf}, substituting in the new gradients:
\begin{equation}\label{leaf2}
    \gamma_{l,ikd,m} = - \frac{(\sum_{n \in l}g^{\beta}_{ikbdn})}{(\sum_{n \in l}h^{\beta}_{ikbdn})}.
\end{equation}

The parallel with gradient descent is now even more striking as all bins included in a terminal node are updated with the same leaf value:
\begin{equation}
    \beta_{ikdb} \mathrel{+}= \nu \gamma_{l,ikd,m}, \quad \forall b \in l,
\end{equation}
where $\nu$ is the learning rate. 

Importantly, at each boosting iteration \emph{only one} feature-degree pair competes for an update, analogous to standard GBDT where features compete via split selection. This results in a diagonal approximation of the Hessian, preserving the scalability of the method. Finally, note that the GAM polynomial form we extract is \emph{exactly} equivalent to standard GBDT ensembles, but it is only valid when feature interactions are restricted. The restriction allows trees split on the same features to be summed. In practice, we implement this by growing trees of depth 1.

\section{ParamBoost} \label{paramboost}

In ParamBoost, we exploit gradient boosting with polynomial trees to compute piecewise cubic shape functions. A key distinction from standard GBDT is that we only keep track of the overall (global) parameters of the piecewise polynomials, rather than storing each individual trees. As such, the memory requirements of ParamBoost is independent of the number of trees\slash boosting iterations in the model, in contrast with typical GBDT models.

\subsection{Continuity and smoothness of shape functions}

Computing leaf values using Equation \ref{leaf2} generally results in discontinuities in the resulting shape functions at leaf boundaries. To ensure continuity, we instead boost \textit{relative} to the split point. Specifically, when considering a split at a specific threshold $u_{b^*}$, we calculate all derivatives using $(x_{kn} - u_{b})^d$ as opposed to $(x^*_{kbn})^d$. This ensures contributions from both sides of the split evaluate to zero at the boundary. With this transformation, the gradient becomes:
\begin{equation} \label{gradient parameter}
    g^{\beta}_{ikbdn} = g_{in}(x_{kn}- u_{b})^d,
\end{equation}
and the Hessian:
\begin{equation} \label{hessian parameter}
\begin{split}
    h^{\beta}_{ikbdn} & = h_{in} (x_{kn} - u_{b})^{2d}.
\end{split}
\end{equation}

This ensures continuity at the split points for all monomials of degree $d \geq 1$. The drawback is that both the first- and second-order derivatives must be precomputed for all candidate split points, increasing the computational and memory cost compared with standard gradient boosting with linear trees. This is the main scalability bottleneck of ParamBoost.

Even with this modification, using constants (i.e. degree-0 terms) at split points results in discontinuities in the shape function. Therefore, to ensure $C^0$ continuity, we can simply disallow splits on monomials of degree $d=0$, instead using a \textit{single global} constant term across all bins. This is achieved with a \textit{quasi-Newton} update with a diagonal approximation of the Hessian, computed by aggregating gradients and Hessians across bins (as in Equation \ref{leaf2}, but with a single leaf $l$ encompassing all bins). Updating this global constant competes against splitting on higher-degree monomials when choosing the best split. 

The result above can be generalised to higher order continuity. For a shape function of maximum degree $D$, we can enforce smoothness $C^{S}$ where $S\in \{0,..,D-1\}$ by disallowing splits on all monomials of $O^S$ or below, instead estimating a single global parameter for these terms across all bins. For instance, enforcing $C^1$ continuity requires excluding all splits of $d=0$ and $d=1$, and $C^2$ continuity requires excluding all splits of $d\leq2$.

\subsection{Monotonicity and curvature constraints}

In many applications, the modeller may wish to ensure monotonicity or convexity of the shape function over the input space, i.e.:
\begin{equation} \label{monotonicity}
    m_kf^{\prime}_{ik}(x) \geq 0 \quad \forall x,
\end{equation}
\begin{equation}
    c_kf^{\prime\prime}_{ik}(x) \geq 0 \quad \forall x,
\end{equation}
where:
\begin{enumerate*}[label=(\roman*)]
    \item $f^{\prime}_{ik}(x)$ and $f^{\prime\prime}_{ik}(x)$ are the first and second derivatives of the shape function of feature~$k$ and class~$i$:
    \item $m_k \in $ $ \{-1,1\}$ specifies decreasing or increasing monotonicity respectively; and
    \item $c_k \in \{-1,1\}$ specifies concavity or convexity respectively.
\end{enumerate*}

To ensure monotonicity, when adding a new tree, all updates $\gamma_{l,i,m}$ must satisfy the following constraints, for degree $d\geq1$:
\begin{equation}
\begin{split}
    m_k d \gamma_{l, ikd,m}(x - u_{l})^{d-1} & \geq m_k f^{\prime}_{ik}(x)
\end{split}
\end{equation}
and for $d=0$:
\begin{equation}
\begin{split}
    m_k(\gamma_{\text{right}, ikd,m} - \gamma_{\text{left}, ikd,m}) & \geq 0.
\end{split}
\end{equation}

For convexity\slash concavity, continuity is required, and so (following the above) terms of $d=0$ are not permitted. The relevant constraints for degree $d \geq 2$ are:
\begin{equation}
\begin{split}
    c_kd(d-1) \gamma_{l, ikd,m}(x - u_l)^{d-2} & \geq c_k f^{\prime\prime}_{ik}(x)
    \end{split}
\end{equation}
and for $d-1$:
\begin{equation}
\begin{split}
    c_k(\gamma_{\text{right}, ikd,m} - \gamma_{\text{left}, ikd,m}) & \geq 0.
\end{split}
\end{equation}

\subsection{Output-specific features}

In a multi-output problem, such as multi-class and multi-label classification and multi-task regression, it is desirable to control which features contribute to which predictive functions. This offers two primary benefits: 
\begin{enumerate*}[label=(\roman*)]
    \item \textbf{improved interpretability} as a feature related to a specific class\slash label will impact only its corresponding shape function; and
    \item (specific to multi-class classification) \textbf{consistency} of softmax probabilities with monotonicity\slash curvature constraints of the shape functions.
\end{enumerate*}

\subsection{Confidence intervals}

As ParamBoost learns parametric, piecewise-cubic shape functions, we can compute formal standard errors and confidence intervals. We exploit the fact that, in maximum likelihood estimation, the variance-covariance matrix of the parameter can be approximated by the inverse of the Hessian matrix. We then make the same assumptions as in gradient boosting (and during training): that the Hessian is diagonal and that the observations are independent and identically distributed. For a parameter $\beta_{ikdb}$ of a class~$i$, feature~$k$, degree~$d$ and bin~$b$, the standard error is:

\begin{equation}\label{standard error}
    \text{SE}(\beta_{ikdb}) = \frac{1}{\sqrt{\sum_{n\in b}h_{in}(x^{*}_{kbn})^{2d}}} 
\end{equation}
where $h_{in}(x^{*}_{kbn})^{2d}$ is the diagonal of the Hessian as defined in in Equation \ref{hessian}. The standard error of the shape function $f_{ik}(x_k)$ is then:

\begin{equation}
    \text{SE}_{\text{pred}, ik} (x_k) = \sum_{d=0}^{D}\sum_{b=1}^{B} \text{SE}(\beta_{ikdb})^2(x^{*}_{kb})^{2d}
\end{equation}

The corresponding 95\% confidence interval is:

\begin{equation}\label{standard error}
\begin{split}
    \text{CI}_{95, ik}(x_k) 
    &= \hat{y}_{ik} \pm 1.96 \cdot  \text{SE}_{\text{pred}, ik} (x_k)
\end{split}
\end{equation}

Note that the assumption of independence between classes and features simplifies this computation greatly, as the Hessian matrix is therefore diagonal, and so inverting it to obtain the variance-covariance matrix is straightforward.

\section{Benchmarks} \label{results}

\subsection{Datasets} 

We benchmark ParamBoost on 11 open-source, real-world tabular datasets covering regression, binary classification, and multi-class classification tasks. 
The datasets are sourced from Kaggle and the UCI Machine Learning Repository \citep{uci}. Table \ref{datasets} provides summary statistics, with brief descriptions of each dataset below. 

\begin{table}[t]
\footnotesize
\centering
\caption{Datasets used for benchmarking. N is the number of instances in the dataset.}
\label{datasets}
\begin{tabular}{lrrrr}
\toprule
\textbf{Dataset} & \textbf{Task} & \textbf{N} & \textbf{\# Features} & \textbf{\# Classes} \\
\midrule
\textit{Housing} & Regression & 506 & 13 & - \\
\textit{Concrete} & Regression & 1030 & 8 & - \\
\textit{Power} & Regression & 9568 & 4 & - \\
\textit{Energy} & Regression & 768 & 8 & - \\
\textit{Protein} & Regression & 45730 & 9 & - \\
\textit{Year} & Regression & 515344 & 90 & - \\
\textit{Fraud} & Binary & 284807 & 30 & 2 \\
\textit{Diabetes} & Binary & 130157 & 175 & 2 \\
\textit{Cover} & Multi-class & 581012 & 54 & 7 \\
\textit{HAR} & Multi-class & 10299 & 561 & 6 \\
\textit{LPMC} & Multi-class & 81086 & 29 & 4 \\
\bottomrule
\end{tabular}

\end{table}

\subsubsection*{Regression tasks}

\textbf{Boston housing} (Housing \citep{harrison1978hedonic}): A regression dataset where the target is the median house price in 1000\$ of owner-occupied homes in the suburbs of Boston.\\
\textbf{Concrete compressive strength} (Concrete \citep{yeh1998modeling}): This dataset is a regression problem where the concrete compressive strength (MPa) is being predicted from the concrete recipe characteristics.\\
\textbf{Combined cycle power plant} (Power \citep{Tfekci2014PredictionOF, combined_cycle_power_plant_294}): A dataset containing hourly averaged ambient observations (such as temperature, relative humidity, ...) and where the goal is to predict the net hourly electrical energy output of a combined gas and steam turbine plant.\\
\textbf{Energy efficiency} (Energy \citep{Tsanas2012AccurateQE, energy_efficiency_242}): A dataset where the heating load and cooling load requirements of a building need to be estimated from its shape and characteristics. We choose to predict the heating load requirement.\\
\textbf{Physicochemical properties of protein tertiary structure} (Protein \citep{physicochemical_properties_of_protein_tertiary_structure_265}): This dataset contains information about the area and spatial distribution of the protein, and the target variable is the size of its residue.\\
\textbf{Year prediction MSD} (Year \citep{year_prediction_msd_203}): A regression dataset where the goal is to predict a song release year from audio features. We follow the author's recommendation to split the dataset in a way that the songs of the same artist cannot be in both the train and test sets.

\subsubsection*{Binary classification task}

\textbf{Credit card fraud detection} (Fraud \citep{inproceedings}): A dataset containing transactions from credit card. The target variable is whether the transaction is fraudulent or not.\\
\textbf{Diabetes detection} (Diabetes \citep{widsdatathon2021, diabetes}): A dataset to determine if the patient has diabetes or not from their health indicators.

\subsubsection*{Multi-class classification task}

\textbf{Forest cover type} (Cover \citep{covertype_31, blackard1999comparative}): A multi-class classification dataset where the goal is to predict the forest cover type of a pixel from features such as elevation, aspect, slope, and soil type.\\
\textbf{Human activity recognition using smartphones} (HAR \citep{anguita2013public}): This dataset is intended to recognise human activities from features obtained by smartphone sensors.\\
\textbf{London passenger mode choice} (LPMC \citep{hillel_recreating_2018}): A transportation dataset where the target variable is the transportation mode used during the trips. The target can be walking, cycling, public transport services, or driving. Additionally, the dataset contains various socio-demographic characteristics and trip characteristics such as travel time (obtained from the Google Maps API) and costs for each mode (walking and cycling are assumed to be free). 

\begin{table*}[htbp]
\centering
\caption{Experiment results on the holdout test set. Values are \textbf{MSE} for regression tasks (housing, concrete, power, energy, protein, and year datasets), \textbf{BCEL} for binary classification tasks (fraud and diabetes datasets), and \textbf{CEL} for multi-class classification tasks (Cover, HAR, and LPMC). All metrics are \emph{lower the better}. For the housing, concrete, power, and energy datasets, experiments are repeated over 10 runs, with the mean and standard deviation reported. The best results for each dataset are highlighted in bold.}
\tiny
\label{tab:results}
\begin{tabular}{lrrrrrrr}
\toprule
 & \textbf{Linear\slash Logistic regression} & \textbf{EBM} & \textbf{NAM} & \textbf{NBM} & \textbf{APLR} & \textbf{MGCV} & \textbf{ParamBoost} \\
\midrule

\multicolumn{8}{c}{\textbf{Regression}} \\
\midrule
\textit{Housing} & 22.017 (± 6.538) & 13.691 (± 4.323) & 19.052 (± 4.215) & 21.013 (± 9.798) & 14.087 (± 3.113) & 21.982 (± 6.407) & \textbf{13.683 (± 4.108)} \\
\textit{Concrete} & 112.939 (± 9.648) & 28.153 (± 3.366) & 52.213 (± 5.415) & 48.165 (± 6.278) & 34.009 (± 3.616) & 113.030 (± 9.512) & \textbf{27.774 (± 3.594)} \\
\textit{Power} & 21.029 (± 0.791) & \textbf{13.386 (± 0.651)} & 20.757 (± 1.240) & 19.263 (± 0.575) & 17.770 (± 0.678) & 21.030 (± 0.786) & 15.859 (± 0.686) \\
\textit{Energy} & 9.103 (± 1.442) & 1.229 (± 0.176) & 8.562 (± 1.435) & 10.662 (± 3.340) & \textbf{1.110 (± 0.196)} & 9.001 (± 1.338) & 1.215 (± 0.143) \\
\textit{Protein} & 27.193 & \textbf{23.494} & 24.224 & 30.901 & 24.718 & 27.186 & 23.584 \\
\textit{Year} & 90.442 & \textbf{85.274} & 86.343 & 94.412 & 85.488 & 90.426 & 85.395 \\

\midrule
\multicolumn{8}{c}{\textbf{Binary Classification}} \\
\midrule
\textit{Fraud} & 0.009 & \textbf{0.003} & \textbf{0.003} & 0.012 & \textbf{0.003} & 0.004 & \textbf{0.003} \\
\textit{Diabetes} & 0.416 & \textbf{0.379} & 0.387 & 0.385 & 0.387 & 0.416 & 0.380 \\

\midrule
\multicolumn{8}{c}{\textbf{Multi-class Classification}} \\
\midrule
\textit{Cover} & 0.951 & 0.599 & 0.613 & 0.618 & -* & -** & \textbf{0.579} \\
\textit{HAR} & 0.052 & 0.030 & 0.061 & 0.170 & 0.057 & -** & \textbf{0.029} \\
\textit{LPMC} & 0.731 & 0.677 & 0.676 & 0.677 & \textbf{0.674} & 1.344 & \textbf{0.674} \\

\bottomrule
\multicolumn{8}{r}{*Time out: more than 48 hours} \\
\multicolumn{8}{r}{**Out of memory}
\end{tabular}
\end{table*}

\subsection{Baseline models}

We compare ParamBoost with state-of-the-art GAMs that:
\begin{enumerate*}[label=(\roman*)]
    \item provide explicit interpretability from input features to predictions, 
    \item support regression, binary, and multi-class classification tasks, and 
    \item have a publicly available, usable implementation.
\end{enumerate*}
Below, we summarise each baseline model and the hyper-parameter search spaces used. 

\textbf{Linear\slash Logistic Regression}: 
used for regression and classification tasks, respectively, to represent the simplest form of a GAM. Both models use L2 regularisation. We use the implementation of the \verb"scikit-learn" library \citep{scikit-learn}, and tune the regularisation strength: uniform in $[10^{-5}, 1]$; keeping all other parameters at default values.

\textbf{EBM}: Explainable Boosting Machine is a tree-based GAM built from shallow boosted trees\citep{nori2019interpretml}. We disable pairwise interactions and tune:
\begin{enumerate*}[label=(\roman*)]
    \item the maximum number of bins in the discrete set: $\{64, 128, 256, 512\}$; 
    \item the minimum samples in a leaf in the discrete set: $\{1, 2, 4, 8, 10, 15, 20\}$;
    \item the minimum sum of hessian in a leaf in the discrete set: $\{0.0001, 0.001, 0.01, 0.1\}$; and 
    \item the learning rate in the interval $[0.001, 0.1]$. 
\end{enumerate*}

\textbf{NAM}: Neural Additive Models is an NN-based GAM that learns a small network per feature. We reimplement the recommended NAM architecture of \citet{radenovic2022neural}, with an MLP with 3 hidden layers of 64, 64 and 32 neurons respectively using the AdamW optimiser \citep{loshchilov2017decoupled}. We tune:
\begin{enumerate*}[label=(\roman*)]
    \item initial learning rate: uniform in $[0.0001, 0.1]$; 
    \item the weight decay: logarithmic in $[10^{-6}, 10^{-3}]$; and 
    \item dropout: $\{0, 0.1, 0.2, 0.3, 0.4, 0.5\}$.
\end{enumerate*}

\textbf{NBM}: Neural Basis Models improve scalability over NAM by learning shared basis functions for each shape function.  We reimplement the recommended NBM architecture of \cite{radenovic2022neural}, restricting feature interactions: 100 shared basis functions learnt with MLPs with 3 hidden layers of 256, 128 and 128 neurons respectively. As with NAM we use AdamW and tune:
\begin{enumerate*}[label=(\roman*)]
    \item initial learning rate: uniform in $[0.0001, 0.1]$, 
    \item weight decay: logarithmic in $[10^{-6}, 10^{-3}]$; and 
    \item dropout: $\{0, 0.1, 0.2, 0.3, 0.4, 0.5\}$.
\end{enumerate*}

\textbf{MGCV}: Mixed GAM Computation Vehicle is a widely used spline-based GAM implementation in R\citep{wood2017generalized}. We use its Python wrapper \verb|PyMGCV|. Features are modelled using \verb|CubicSpline| bases with 10 knots (or fewer if the number of unique feature values is small). We use linear terms for dummy or binary features. We tune:
\begin{enumerate*}[label=(\roman*)]
    \item using fixed degrees of freedom vs. penalised splines (\verb|fx| hyperparameter), and 
    \item whether to include a penalty on the null space (\verb|shrinkage| hyperparameter).
\end{enumerate*}

\textbf{APLR}: Automatic Piecewise-Linear Regression \citep{von2024automatic} is a piecewise linear GAM implemented in the \verb|interpret| library. We tune:
\begin{enumerate*}[label=(\roman*)]
    \item learning rate: uniform in$[0.001, 1]$;
    \item L2 penalty: $\{0.0001, 0.001, 0.01, 0.01, $ $1\}$; and 
    \item the minimum of observations per split: $\{1, 2, 4, $ $8, 10, 15, 20\}$.
\end{enumerate*}
Default values are used for all remaining parameters.

\subsection{Implementation details}

We implement ParamBoost with the \verb|JAX| Python library \citep{jax2018github}. For the terms of degree $d > 0$, we are required to multiply functional space gradients with features, as defined in Equation \ref{gradient parameter} and Equation \ref{hessian parameter}. To compute them efficiently, we need to precompute the feature values for all potential split points and order. This is the main bottleneck of ParamBoost in terms of memory requirements, and, to keep the model scalable, we use a different number of bins for monomials of degree $d=0$ and monomials of degree $d>0$. We find experimentally that ParamBoost is less sensitive to hyperparameters, therefore we run the model with default parameters, that is: \begin{enumerate*}[label=(\roman*)]
    \item learning rate: $0.1$;
    \item L1 penalty: $0.001$
    \item L2 penalty: $0.01$; 
    \item Minimum number of data in a leaf: $10$; 
    \item Max number of bins for terms of degree $d=0$: $256$; and
    \item Max number of bins for terms of degree $d>0$: $20$;
\end{enumerate*} 
Since we are using trees of depth 1, we do not use any other regularisation hyperparameters for tree building. For the benchmarks, we use the most flexible version of ParamBoost, that is, cubic polynomials with no required continuity or differentiability. ParamBoost will be released as a Python package upon acceptance of the paper.

For all datasets, we split the data into 70 \% for training, 10 \% for validation, and 20 \% for testing. We randomly search for optimal hyperparameters (except for ParamBoost) and choose them on the basis of the model's performance on the validation set. For ParamBoost, APLR, and EBM, we train the model for a maximum of 25000 iterations, and we stop the training if the validation loss does not improve during 100 iterations. Note that for the last two models, it is not possible to choose the instances used for validation, so early stopping is done with cross-validation on the training set. For NAM and NBM, we train the models for a maximum of 200 epochs and stop the training if the validation loss does not improve during 10 epochs. We train MGCV and Linear\slash Logistic Regression with the default method. For \textit{housing}, \textit{concrete}, \textit{power} and \textit{energy} datasets, we repeat the experiment 10 times to mitigate the randomness due to the train–validation–test split, and search for 50 sets of hyperparameters. For all other datasets, we do a hyperparameter search with 15 trials. We train ParamBoost, NAM, and NBM on a single A5000 GPU with 24GB of memory, and linear\slash logistic, EBM, MGCV, and APLR on the CPU. We use the Mean-Squared Error (MSE), Binary Cross-Entropy Loss (BCEL), and Cross-Entropy Loss (CEL) as loss functions for regression, binary, and multi-class classification tasks.

\section{Results} \label{benchmarks}

The experiment results are shown in Table \ref{tab:results}. Overall, we observe that ParamBoost, EBM, and APLR consistently outperform other GAMs, with ParamBoost showcasing the best performance overall. We note that ParamBoost performs especially well on classification tasks, and on small regression datasets, while EBM showcase better performances on bigger regression datasets and binary classification datasets. The splines or NN-based GAMs mostly outperform linear or logistic regression, although not strictly, depending on the dataset. Table \ref{tab:time} in Appendix \ref{computational time} summarises computational times for all models and datasets. Linear or logistic regression is significantly faster than all other models. Among GAMs, NAM is the fastest model on small datasets, whereas EBM and NBM are the most efficient models on larger datasets. ParamBoost is slower than other models because:
\begin{enumerate*}[label=(\roman*)]
    \item since the derivatives are computed in the parameter space, they need to be recomputed for each potential threshold at each iteration;
    \item there are more potential split points due to the cubic nature of the model; and
    \item \verb|JAX| includes a compilation time.
\end{enumerate*}
However, we note that ParamBoost is significantly faster than APLR, the only other model in this benchmark that has trees with degree $d>0$, and that, since we do not perform a hyperparameter search for ParamBoost, the training time is overall faster than for all other models.

\begin{figure*}[!htbp]
    \centering
    \begin{subfigure}[b]{.24\textwidth}
        \centering
        \includegraphics[width=\textwidth]{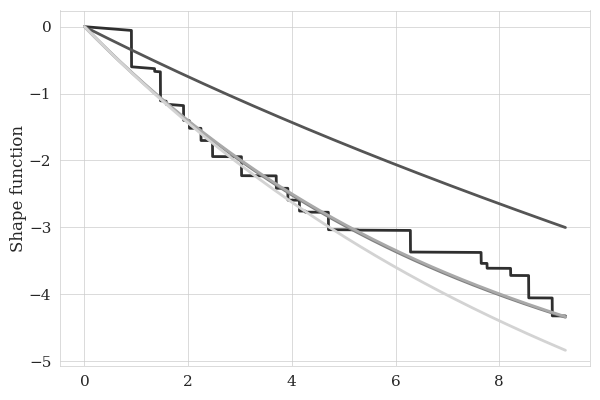}
        \caption{Walking travel time.}
        \label{fig:walking}
    \end{subfigure}
    \begin{subfigure}[b]{.24\textwidth}
        \includegraphics[width=\textwidth]{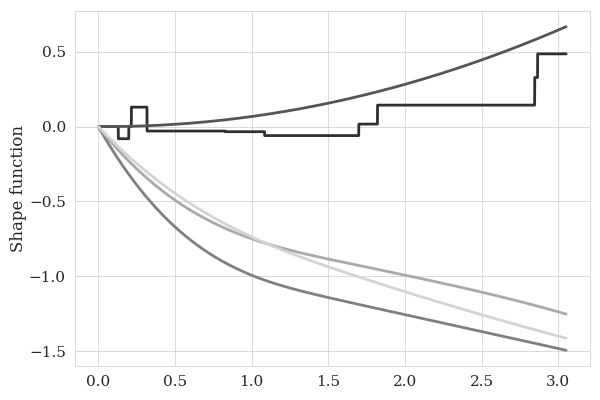}
        \caption{Cycling travel time}
        \label{fig:cycling}
    \end{subfigure}
    \begin{subfigure}[b]{.24\textwidth}
        \centering
        \includegraphics[width=\textwidth]{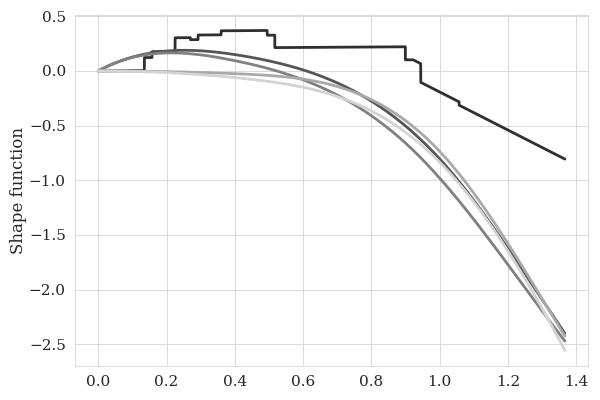}
        \caption{Rail travel time.}
        \label{fig:rail}
    \end{subfigure}
    \begin{subfigure}[b]{.24\textwidth}
        \includegraphics[width=\textwidth]{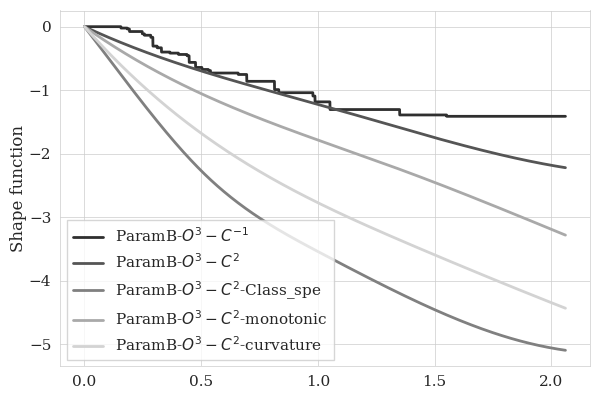}
        \caption{Driving travel time.}
        \label{fig:driving}
    \end{subfigure}
    \caption{Travel time shape functions.}
    \label{fig:case_study_all_shapes}
\end{figure*}

\begin{figure*}[!htbp]
    \centering
    \begin{subfigure}[b]{.24\textwidth}
        \centering
        \includegraphics[width=\textwidth]{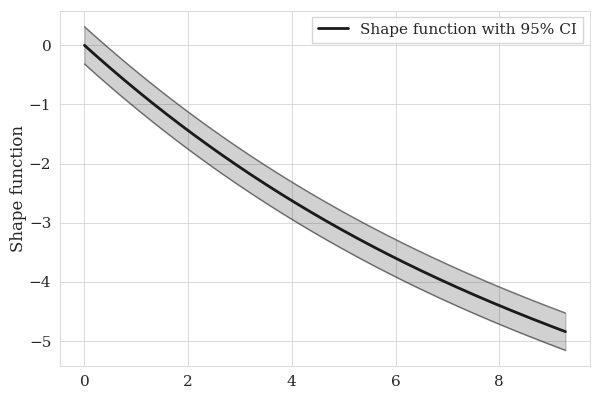}
        \caption{Walking travel time.}
        \label{fig:walking2}
    \end{subfigure}
    \begin{subfigure}[b]{.24\textwidth}
        \includegraphics[width=\textwidth]{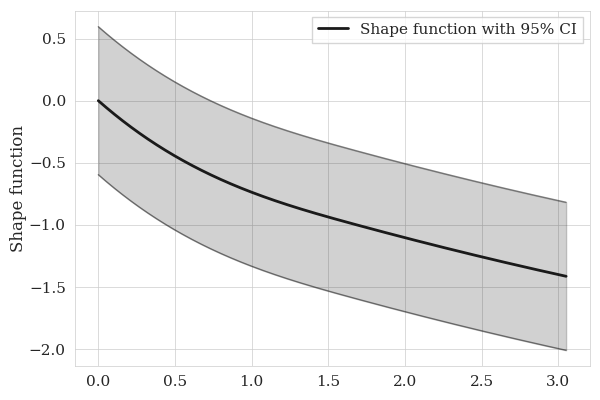}
        \caption{Cycling travel time}
        \label{fig:cycling2}
    \end{subfigure}
    \begin{subfigure}[b]{.24\textwidth}
        \centering
        \includegraphics[width=\textwidth]{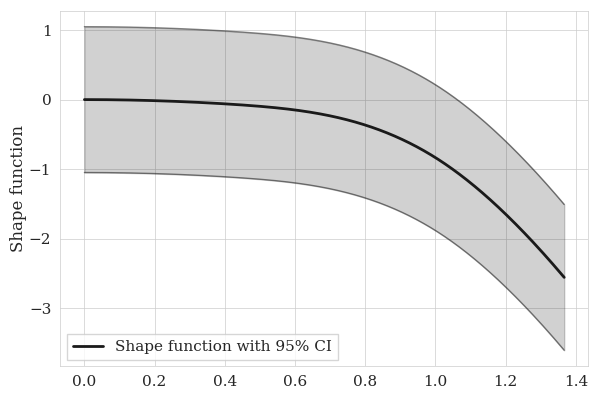}
        \caption{Rail travel time.}
        \label{fig:rail2}
    \end{subfigure}
    \begin{subfigure}[b]{.24\textwidth}
        \includegraphics[width=\textwidth]{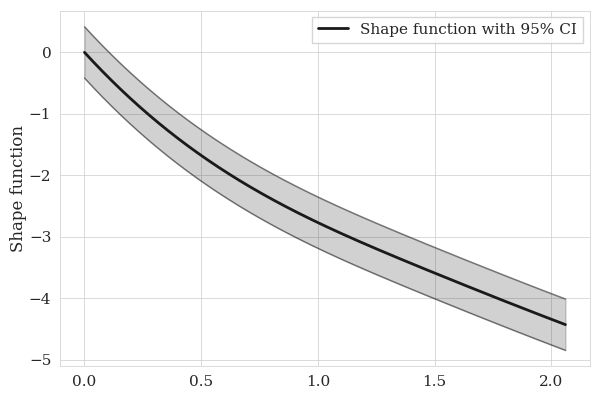}
        \caption{Driving travel time.}
        \label{fig:driving2}
    \end{subfigure}
    \caption{Travel time shape functions with 95\% confidence interval}
    \label{fig:case_study}
\end{figure*}

\subsection{Interpretability case study: LPMC}

While Section \ref{benchmarks} compares the model based purely on its predictive performance, extracting parametric uncertainties and marginal effects is key for the model acceptance by practitioners. We use the LPMC dataset from the benchmarks section to conduct an evaluation of the interpretability and trustworthiness of ParamBoost. 

To understand the effect of the constraints on the model, we start from the flexible model of Section \ref{benchmarks}, then sequentially add differentiability constraints, class-specific feature constraints, monotonicity constraints, and finally curvature constraints. For class-specific features, each feature impacts only its respective class predictive function (e.g., walking travel time impacts only the walking predictive function). Using domain knowledge, we apply negative monotonicity and convexity constraints to all cost, travel time, transfer and congestion rate features. This means that an increase in these features is usually perceived negatively, and worse when their values are small than when they are bigger (e.g. an increase of 1 minute in travel time for a 10-minute trip is perceived more negatively than an increase of 1 minute in travel time for a 3-hour trip). Interestingly, we have found empirically that the convex assumption was not justified for bus and rail travel time; therefore, we apply a concave constraint to the shape functions of these two features.

The cross-entropy loss for the five models is shown in Table \ref{tab:ablation}. We observe that adding constraints marginally decreases the model predictive performance. More specifically, in this case study, monotonicity constraints seem to have the worst effect on predictive performance, unless coupled with curvature constraints. Overall, the final model with all constraints is a model with more meaningful shape functions than the original one, at a modest cost of predictive performance. This highlights the modest trade-off that is required for better interpretability.

\begin{table}[htbp]
\small
\centering
\caption{A case study where each model successively has more constraints. We first add differentiability constraints, then class-specific features, monotonicity, and curvature constraints. We compare the models on their cross-entropy loss.}
\label{tab:ablation}
\begin{tabular}{lr}
\toprule
 & \textbf{CEL} \\
\midrule
\textit{ParamBoost-$O^3-C^{-1}$} & 0.6734 \\
\textit{ParamBoost-$O^3-C^2$} & 0.6749 \\
\textit{ParamBoost-$O^3-C^2$-ClassSpec} & 0.6744 \\
\textit{ParamBoost-$O^3-C^2$-MonoCons} & 0.6792 \\
\textit{ParamBoost-$O^3-C^2$-CurvCons} & 0.6757 \\
\bottomrule
\end{tabular}
\end{table}

We can then observe the $C^2$ piecewise cubic shape functions learnt from the data. As an example, we provide the shape functions of the walking, cycling, rail and driving travel times, in Figure \ref{fig:case_study_all_shapes}, for all models. We see that, apart from the cycling travel time, the shape functions are fairly similar to those for the most flexible, but with more guarantee of interpretability. For cycling travel time, the constraints ensure that the shape functions are consistent with behavioural theory, as an increase in travel time should decrease the attractiveness of an alternative.

Finally, we can also draw 95\% confidence intervals (CI). Figure \ref{fig:case_study} shows the shape functions of the model with $C^2$, class-specific features, monotonicity, and curvature constraints with 95\% CI.

\section{Conclusion} \label{conclusions}

To conclude, we introduce ParamBoost, a model that learns piecewise cubic shape functions that are up to $C^2$, such that elasticities can be extracted for any input feature. With ParamBoost, we also provide the practitioner with control over the model interpretability. That means that it is possible to select constraints at the discretion of the modeller. This provide the modeller with keys to incorporate expert knowledge. The trade-off between interpretability and predictive performance can be assessed, with often having a much simpler model performing as well as or even better than a much more complex one. Finally, this unlocks new applications of interpretable ML models in fields where continuous derivatives are crucially needed.

\appendix

\section{Training and validation results on the benchmarks}

Table \ref{tab:results_train} and Table \ref{tab:results_val} summarise the results on the train and validation sets for all models and all datasets.

\begin{table*}[htbp]
\centering
\caption{Experiment results on the train set. We compare the models with their \textbf{MSE} for regression tasks (housing, concrete, power, energy, protein, and year datasets), \textbf{BCEL} for binary classification tasks (fraud and diabetes datasets), and \textbf{CEL} for multi-class classification tasks (Cover, HAR, and LPMC). All metrics are \emph{lower the better}. For the housing, concrete, power, and energy datasets, results are the mean over 10 runs and their corresponding standard deviation.}
\tiny
\label{tab:results_train}
\begin{tabular}{lrrrrrrr}
\toprule
 & \textbf{Linear\slash Logistic regression} & \textbf{EBM} & \textbf{NAM} & \textbf{NBM} & \textbf{APLR} & \textbf{MGCV} & \textbf{ParamBoost} \\
\midrule

\multicolumn{8}{c}{\textbf{Regression}} \\
\midrule
Housing & 22.344 (± 1.591) & 8.700 (± 3.873) & 17.816 (± 2.659) & 19.080 (± 4.191) & 11.575 (± 1.438) & 22.149 (± 1.483) & 8.310 (± 2.674) \\
Concrete & 106.516 (± 2.827) & 14.506 (± 1.806) & 48.906 (± 3.300) & 44.671 (± 6.431) & 29.757 (± 2.297) & 106.319 (± 2.802) & 17.237 (± 4.026) \\
Power & 20.725 (± 0.236) & 10.814 (± 0.851) & 20.101 (± 0.669) & 18.769 (± 0.413) & 17.094 (± 0.260) & 20.724 (± 0.236) & 14.251 (± 0.483) \\
Energy & 8.578 (± 0.429) & 0.993 (± 0.040) & 8.379 (± 0.799) & 10.156 (± 2.495) & 0.910 (± 0.045) & 8.473 (± 0.377) & 0.997 (± 0.034) \\
Protein & 26.817 & 22.353 & 23.903 & 30.362 & 24.338 & 26.815 & 22.529 \\
Year & 91.013 & 83.620 & 85.666 & 93.787 & 84.447 & 91.009 & 83.462 \\

\midrule
\multicolumn{8}{c}{\textbf{Binary Classification}} \\
\midrule
Fraud & 0.009 & 0.002 & 0.002 & 0.013 & 0.003 & 0.004 & 0.003 \\
Diabetes & 0.414 & 0.371 & 0.382 & 0.379 & 0.384 & 0.414 & 0.371 \\

\midrule
\multicolumn{8}{c}{\textbf{Multi-class Classification}} \\
\midrule
Cover & 0.955 & 0.595 & 0.611 & 0.617 & -* & -** & 0.572 \\
HAR & 0.028 & 0.005 & 0.022 & 0.135 & 0.041 & -** & 0.005 \\
LPMC & 0.708 & 0.624 & 0.649 & 0.652 & 0.650 & 1.354 & 0.636 \\

\bottomrule
\multicolumn{8}{r}{*Time out: more than 48 hours} \\
\multicolumn{8}{r}{**Out of memory}
\end{tabular}
\end{table*}

\begin{table*}[htbp]
\centering
\caption{Experiment results on the validation set. We compare the models with their \textbf{MSE} for regression tasks (housing, concrete, power, energy, protein, and year datasets), \textbf{BCEL} for binary classification tasks (fraud and diabetes datasets), and \textbf{CEL} for multi-class classification tasks (Cover, HAR, and LPMC). All metrics are \emph{lower the better}. For the housing, concrete, power, and energy datasets, results are the mean over 10 runs and their corresponding standard deviation.}
\tiny
\label{tab:results_val}
\begin{tabular}{lrrrrrrr}
\toprule
 & \textbf{Linear\slash Logistic regression} & \textbf{EBM} & \textbf{NAM} & \textbf{NBM} & \textbf{APLR} & \textbf{MGCV} & \textbf{ParamBoost} \\
\midrule

\multicolumn{8}{c}{\textbf{Regression}} \\
\midrule
Housing & 20.899 (± 6.442) & 10.473 (± 2.997) & 14.713 (± 6.082) & 17.981 (± 6.104) & 11.479 (± 4.314) & 23.006 (± 6.831) & 12.045 (± 4.391) \\
Concrete & 103.985 (± 14.585) & 24.797 (± 6.127) & 50.568 (± 10.201) & 44.914 (± 9.001) & 35.597 (± 6.398) & 106.965 (± 14.900) & 27.846 (± 8.078) \\
Power & 20.574 (± 1.117) & 12.785 (± 1.216) & 20.097 (± 1.635) & 18.730 (± 1.159) & 17.286 (± 1.128) & 20.594 (± 1.119) & 15.391 (± 1.323) \\
Energy & 8.073 (± 1.632) & 1.069 (± 0.355) & 7.664 (± 1.345) & 9.055 (± 3.219) & 0.925 (± 0.310) & 8.287 (± 1.595) & 1.064 (± 0.359) \\
Protein & 26.598 & 22.752 & 23.785 & 30.197 & 24.133 & 26.615 & 22.800 \\
Year & 92.963 & 86.647 & 88.220 & 96.449 & 86.814 & 93.020 & 86.697 \\

\midrule
\multicolumn{8}{c}{\textbf{Binary Classification}} \\
\midrule
Fraud & 0.008 & 0.002 & 0.002 & 0.012 & 0.003 & 0.004 & 0.003 \\
Diabetes & 0.413 & 0.382 & 0.387 & 0.389 & 0.389 & 0.415 & 0.382 \\

\midrule
\multicolumn{8}{c}{\textbf{Multi-class Classification}} \\
\midrule
Cover & 0.953 & 0.599 & 0.613 & 0.617 & -* & -** & 0.579 \\
HAR & 0.028 & 0.035 & 0.069 & 0.181 & 0.062 & -** & 0.034 \\
LPMC & 0.716 & 0.653 & 0.661 & 0.662 & 0.649 & 1.332 & 0.650 \\

\bottomrule
\multicolumn{8}{r}{*Time out: more than 48 hours} \\
\multicolumn{8}{r}{**Out of memory}
\end{tabular}
\end{table*}

\section{Computational time}\label{computational time}

Table \ref{tab:time} summarises the best training iteration for all models and all datasets.

\begin{table*}[bp]
\centering
\caption{Computational time in seconds for the best training iteration.}
\tiny
\label{tab:time}
\begin{tabular}{lrrrrrrr}
\toprule
 & \textbf{Linear\slash Logistic regression} & \textbf{EBM} & \textbf{NAM} & \textbf{NBM} & \textbf{APLR} & \textbf{MGCV} & \textbf{ParamBoost*} \\
\midrule

\multicolumn{8}{c}{\textbf{Regression}} \\
\midrule
Housing & 0.001 (± 0.000) & 0.271 (± 0.255) & 0.352 (± 0.147) & 0.453 (± 0.171) & 0.978 (± 0.856) & 0.016 (± 0.001) & 2.338 (± 3.562) \\
Concrete & 0.001 (± 0.000) & 1.163 (± 1.207) & 0.414 (± 0.145) & 0.507 (± 0.125) & 7.125 (± 2.788) & 0.017 (± 0.001) & 3.368 (± 3.293) \\
Power & 0.001 (± 0.000) & 5.156 (± 0.921) & 0.915 (± 0.351) & 1.355 (± 0.643) & 19.485 (± 7.085) & 0.053 (± 0.014) & 6.492 (± 4.575) \\
Energy & 0.002 (± 0.001) & 1.005 (± 1.094) & 0.399 (± 0.133) & 0.407 (± 0.130) & 0.979 (± 0.134) & 0.013 (± 0.001) & 1.556 (± 2.611) \\
Protein & 0.004 & 2.298 & 9.292 & 5.740 & 91.163 & 0.742 & 25.829 \\
Year & 0.308 & 66.343 & 230.872 & 255.686 & 2133.798 & 121.807 & 611.758 \\

\midrule
\multicolumn{8}{c}{\textbf{Binary Classification}} \\
\midrule
Fraud & 1.117 & 10.429 & 67.630 & 15.004 & 1005.472 & 10.027 & 22.359 \\
Diabetes & 1.064 & 7.660 & 82.196 & 53.178 & 310.639 & 40.006 & 207.932 \\

\midrule
\multicolumn{8}{c}{\textbf{Multi-class Classification}} \\
\midrule
Cover & 8.728 & 371.311 & 329.540 & 135.860 & - & - & 1078.351 \\
HAR & 0.720 & 14.163 & 60.445 & 10.421 & 903.500 & - & 93.031 \\
LPMC & 0.674 & 13.207 & 15.862 & 7.443 & 494.494 & 11.891 & 30.636 \\

\bottomrule
\multicolumn{8}{r}{\textit{*Full training time}}\\
\end{tabular}
\end{table*}

\printcredits

\section*{Declaration of competing interest}
The authors declare that they have no known competing financial interests or personal relationships that could have appeared to influence the work reported in this paper.

\section*{Acknowledgments}
This research is supported by a Chadwick PhD Scholarship awarded to Nicolas Salvadé by the UCL Department of Civil, Environmental, and Geomatic Engineering.

\bibliographystyle{cas-model2-names}

\bibliography{cas-refs}



\end{document}